\documentclass[10pt,twocolumn,letterpaper]{article}

\usepackage[pagenumbers]{cvpr} % To force page numbers, e.g. for an arXiv version

\usepackage[dvipsnames]{xcolor}

\definecolor{cvprblue}{rgb}{0.21,0.49,0.74}
\usepackage[pagebackref,breaklinks,colorlinks,citecolor=cvprblue]{hyperref}
\usepackage{pifont}
\usepackage{multirow}
\usepackage{array}
\newcommand*\rot{\rotatebox{90}}

\def\paperID{****} % *** Enter the Paper ID here
\def\confName{CVPR}
\def\confYear{2024}

\title{Synchronizing Vision and Language: Bidirectional Token-Masking AutoEncoder for Referring Image Segmentation}

\author{Minhyeok Lee$^{1}$ \quad
	Dogyoon Lee$^{1}$ \quad
	Jungho Lee$^{1}$ \quad
	Suhwan Cho$^{1}$ \\
	Heeseung Choi$^{2}$ \quad
	Ig-Jae Kim$^{2}$ \quad
	Sangyoun Lee$^{1,2}$ \\
	\vspace{-0.1cm}
	$^{1}$Yonsei University\\
	$^{2}$Korea Institute of Science and Technology (KIST)\\
	{\tt\small \{hydragon516, nemotio, 2015142131, chosuhwan, syleee\}@yonsei.ac.kr}\\
	{\tt\small \{hschoi, drjay\}@kist.re.kr}
}

\begin{document}
\maketitle
\begin{abstract}
	Referring Image Segmentation (RIS) aims to segment target objects expressed in natural language within a scene at the pixel level. Various recent RIS models have achieved state-of-the-art performance by generating contextual tokens to model multimodal features from pretrained encoders and effectively fusing them using transformer-based cross-modal attention. While these methods match language features with image features to effectively identify likely target objects, they often struggle to correctly understand contextual information in complex and ambiguous sentences and scenes. To address this issue, we propose a novel bidirectional token-masking autoencoder (BTMAE) inspired by the masked autoencoder (MAE). The proposed model learns the context of image-to-language and language-to-image by reconstructing missing features in both image and language features at the token level. In other words, this approach involves mutually complementing across the features of images and language, with a focus on enabling the network to understand interconnected deep contextual information between the two modalities. This learning method enhances the robustness of RIS performance in complex sentences and scenes. Our BTMAE achieves state-of-the-art performance on three popular datasets, and we demonstrate the effectiveness of the proposed method through various ablation studies.
\end{abstract} 

\section{Introduction}
\label{intro}
Referring Image Segmentation (RIS) aims to segment the target object indicated by natural language expressions from a single RGB image. Recent advancements in large language models have led to experimental research, actively attempting to integrate computer vision models with natural language models. In particular, RIS has attracted attention due to its substantial potential in image editing and generation~\cite{hertz2022prompt, song2023clipvg, crowson2022vqgan} based on natural language, as well as in the field of human-robot interaction~\cite{wang2019reinforced}.

\begin{figure}[t]
	\setlength{\belowcaptionskip}{-24pt}
	\begin{center}
		\includegraphics[width=\linewidth]{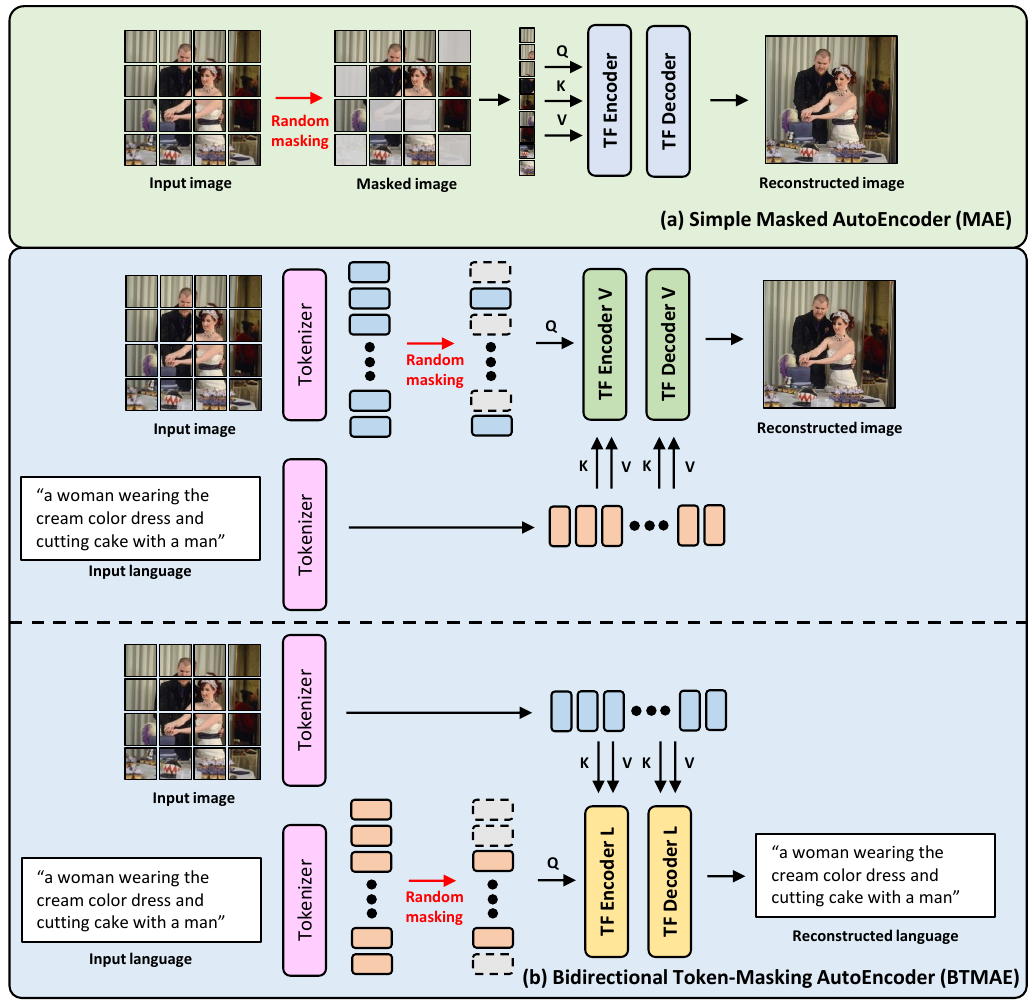}
		\vspace{-0.7cm}
		\caption{(a) The structure of the traditional MAE. (b) The structure of the proposed BTMAE. BTMAE reconstructs the original image from masked visual tokens and original language tokens, and it also reconstructs the original language sentence from the original visual tokens and masked language tokens.  With this, BTMAE effectively models high-dimensional relationships among each multimodal token.}
		\label{fig:intro}
	\end{center}
\end{figure}

Most recent RIS works~\cite{ding2021vision, wang2022cris, yang2022lavt, tang2023contrastive, liu2023gres, kim2022restr} aim to tokenize visual and text features based on powerful transformer-based encoders and model multimodal relationships using cross-attention. These approaches effectively match target vision features from natural language descriptions with the help of similarity-based attachment mechanisms. However, these approaches often produce incorrect visual-language relationship modeling in challenging scenes. For example, in the case of multiple foreground objects and complex backgrounds, esoteric modifier expressions, or complex sentences, these methods fail to model the higher-order visual-language relationships and predict incorrect target masks. In particular, in a method based on indiscriminate token integration, a large number of similar objects and unnecessary background descriptions may transmit unnecessary multimodal information to the decoder, hindering accurate mask generation. Some works~\cite{wang2022cris, yang2022lavt, tang2023contrastive} try to solve these problems by extracting tokens with rich multimodal contextual information and highlighting the connectivity between them using methods such as contrastive learning, visual-language transformer fintuning, and mutimodal embedding space modeling. However, these methods still struggle with modeling challenging high-dimensional visual-language contexts. 

\begin{figure*}[t]
	\setlength{\belowcaptionskip}{-24pt}
	\begin{center}
		\includegraphics[width=\linewidth]{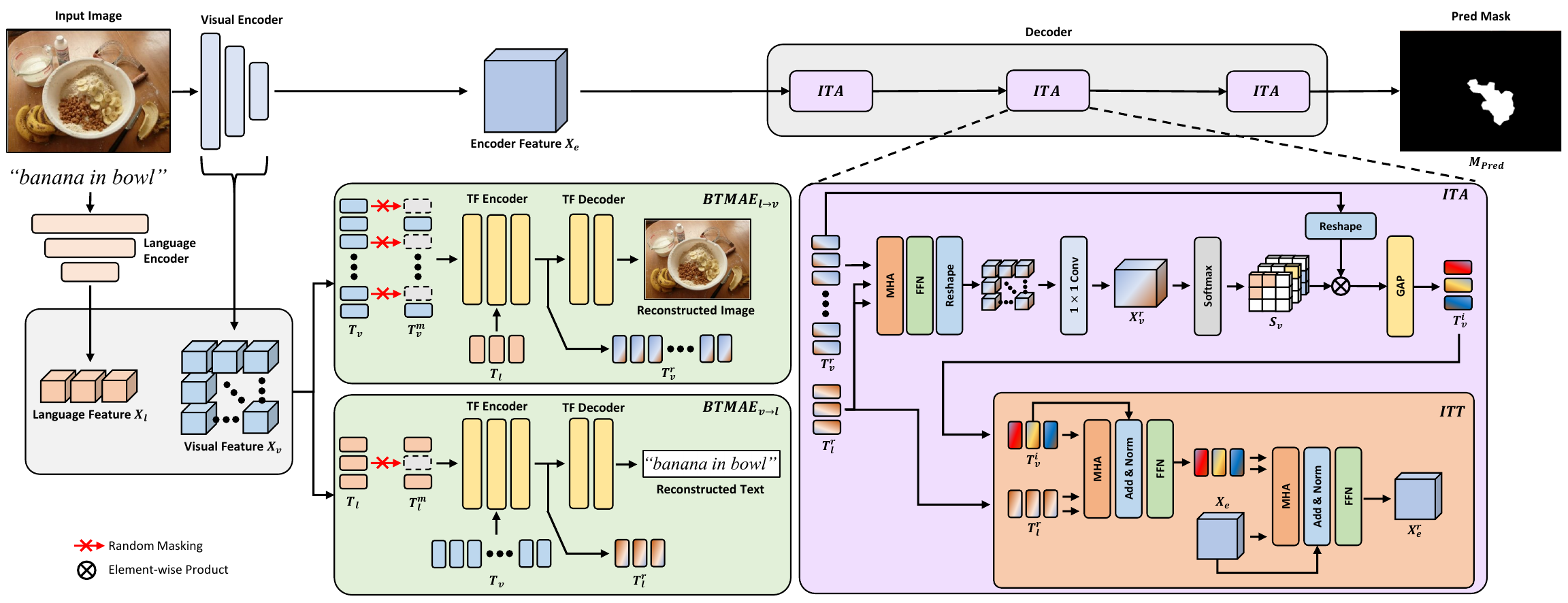}
		\vspace{-0.7cm}
		\caption{Overall structure of the proposed model. The proposed model consists of a visual and a language encoder for modality feature extraction, two types of BTMAEs for cross-modal relation modeling, and ITAs for impact token extraction.}
		\label{fig:main}
	\end{center}
\end{figure*}

To address these issues, we focus on the self-feature modeling ability of the Masked Autoencoder (MAE)~\cite{he2022masked}. The well-known pretraining method, MAE, involves masking images at the patch level and using a transformer-based autoencoder to reconstruct the original image. Therefore, MAE enhances the generalization and inference performance of downstream task models by self learning how to reconstruct overall visual context information based on sparse visual cues. To take advantage of the MAE, some works~\cite{wang2023image, lin2023smaug} improve visual-language context feature modeling performance by applying the MAE method to each modality of a visual-language pretraining task. However, there are several challenges that make it difficult to apply MAE to the RIS task. First, it is widely recognized that MAE-based pre-training methods require large datasets, such as ImageNet~\cite{deng2009imagenet}, to effectively generalize the model. However, the RIS dataset requires input images, segmentation masks for target objects, and natural language representations for the target objects. As a result, the dataset is relatively small, making it challenging to generalize with the MAE approach. While it is possible to train MAE using large-scale image-text datasets proposed by CLIP~\cite{radford2021learning} or SAM~\cite{kirillov2023segment}, excluding the target object's segmentation mask, it becomes difficult to fairly compare these additional datasets with the traditional RIS task approach. Furthermore, some recent approaches~\cite{lee2023weakly, liu2023referring, kim2023shatter} employ weakly-supervised methods in RIS tasks to address the issue of limited datasets, but their performance is considerably lower than that of fully-supervised methods. Second, traditional MAE based on single-modality self-attention cannot learn the relationships between multiple modalities. To address this issue,~\cite{wang2023image} applies one multimodal MAE to image and text information simultaneously to model intermodal information between visual languages. However, this simultaneous masking of image and text is difficult to apply to RIS tasks because it can result in missing target object information for all modalities. In other words, when both the image region and the text words of the target object are removed, it results in a complete loss of information for generating visual-language relationships.

In this paper, we propose a novel bidirectional token-masking token autoencoder (BTMAE) to address the challenges of the RIS task and the drawbacks of MAE. Figure~\ref{fig:intro} illustrates the structural differences between the traditional MAE and the proposed BTMAE. As shown in the figure, BTMAE embeds a input image and text into tokens and randomly masks these tokens. Unlike MAE, which masks at the patch level, tokens encoded by the tokenizer in BTMAE establish attention-based self-correlations within each modality. As a result, the features encoded in BTMAE can contain more contextual information than MAE. This structure, consequently, helps maintain the generalization performance of MAE on the previously mentioned limited RIS dataset. Furthermore, as shown in Figure~\ref{fig:intro}, we divide BTMAE into two streams to model high-dimensional relational information between modalities: one stream reconstructs missing images from intact text, and the other stream reconstructs missing text from intact images. In other words, each stream omits information from one of the modalities, and the unaltered modality is used to restore this missing modality. This bidirectional two-stream structure effectively models cross-modal context features from language to visual and from visual to language directions, synchronizing the image and text. Furthermore, to maintain the robustness of our model on challenging datasets that include unnecessary words and complex backgrounds, we propose a decoding layer called Impact Token Attention (ITA). ITA samples essential impact tokens for generating the correct masks from the visual and language tokens produced by BTMAEs through self-attention. Using this process, the proposed model removes as much noise as possible from ambiguous sentences and multiple objects and generates an accurate predict mask.

Our method was evaluated on three widely-used datasets: RefCOCO~\cite{yu2016modeling}, RefCOCO+~\cite{yu2016modeling}, GRef~\cite{mao2016generation, nagaraja2016modeling}. These datasets consist of diverse and challenging scenarios, and our proposed model achieves state-of-the-art performance on all three. Furthermore, through various ablation studies, we have demonstrated the effectiveness of our model and shown that it can achieve robust RIS even in challenging scenes.

Our main contributions can be summarized as follows:
\begin{itemize}
	\item We propose BTMAE, inspired by MAE, to address the shortcomings of previous RIS models. Unlike traditional MAE, BTMAE can capture multi-modal context information and effectively enhance the model's performance even with a non-large datasetst.
	
	\item To eliminate the impact of irrelevant information in complex sentences and crowded foreground scenes, we introduce the ITA module. ITA samples multi-modal tokens generated by BTMAE to create essential impact tokens.
	
	\item The proposed method achieves state-of-the-art performance on three popular datasets and demonstrates robustness in challenging scenes through various ablation studies.
\end{itemize}

\section{Related Work}
\noindent
\textbf{Referring Image Segmentation.}
RIS is a task that involves segmenting a specific object or region in an image referred to by a natural language expression, allowing for precise localization of the referred object. Unlike traditional methods~\cite{hu2016segmentation, liu2017recurrent, yu2018mattnet, li2018referring, margffoy2018dynamic, shi2018key} that map image feature maps to a language feature space through simple convolutional operations, recent approaches~\cite{ding2021vision, wang2022cris, yang2022lavt, tang2023contrastive, liu2023gres, kim2022restr} have gained prominence by employing the attention mechanism of transformers to achieve higher-dimensional visual-language feature fusion. First, the works in~\cite{ding2021vision, wang2022cris, yang2022lavt} introduce transformer-based architecture to efficiently locate the region in an image that corresponds to a given language expression. These similarity-based methods work well for short sentences and uncomplicated scenes, but often fail for sentences containing unnecessary descriptions and complex backgrounds. To solve this problem,~\cite{tang2023contrastive, liu2023gres, kim2022restr} focuses on tokenizing objects and texts into independent features and improving the fusing and embedding capabilities of these two modalities. In other words, they aim to extract image tokens and language tokens from the encoder and model their cross-modal dependencies based on cross-attention. However, these approaches still rely on the association between image and language features, making it challenging to clearly understand the contextual relationships between in complex sentences and scenes. In contrast, the proposed BTMAE is inspired by MAE in unsupervised learning and can model cross-modal context information, which is essential in RIS tasks.

\noindent
\textbf{Masked Autoencoder.} The MAE~\cite{he2022masked} is a type of neural network architecture that learns to reconstruct its input data while selectively ignoring or masking certain parts of the input, typically used for feature learning and dimensionality reduction. Masked autoencoders can be successful in various computer vision tasks~\cite{wang2022contrastmask, xie2023maester, wu2023dropmae, voleti2022mcvd}, such as image segmentation, tracking, and generation, because they can capture and represent meaningful features in images by learning to reconstruct them while ignoring noisy or less relevant information. The proposed BTMAE can effectively enhance RIS performance by leveraging the feature representation modeling capability of MAE to learn complex contextual relationships between modalities in multimodal spaces.

\begin{figure*}[t]
	\setlength{\belowcaptionskip}{-24pt}
	\begin{center}
		\includegraphics[width=\linewidth]{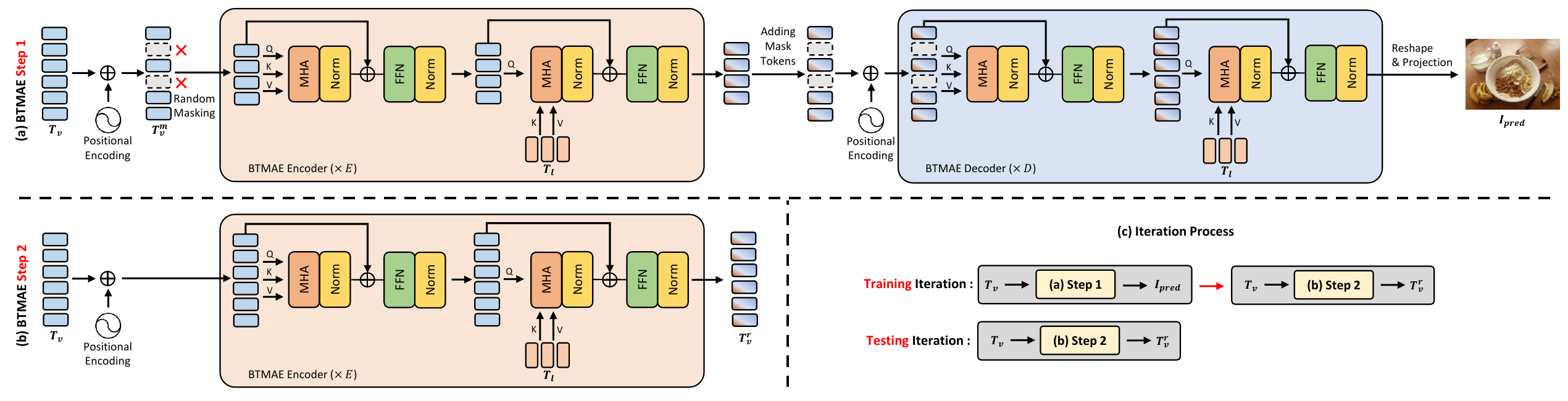}
		\vspace{-0.7cm}
		\caption{The structure of the proposed BTMAE's step (a) and step (b), as well as the processes in the training and testing phases (c). While (a) and (b) are only represented for $BTMAE_{\left(l \rightarrow v \right)}$, $BTMAE_{\left(v \rightarrow l \right)}$ also has the same model architecture. In the model training iteration, steps (a) and (b) are executed sequentially, but in the testing iteration, only (b) is executed.}
		\label{fig:3mae}
	\end{center}
\end{figure*}

\section{Proposed Approach}
\subsection{Overall Architecture}
Figure~\ref{fig:main} shows the overall structure of the proposed model. In the same manner as previous RIS tasks~\cite{ding2021vision, wang2022cris, yang2022lavt}, our model takes as input a single RGB image and a text description of the target object. The proposed model consists of visual and language encoders for feature extraction, two BTMAEs for enhancing context information. Furthermore, the ITA decoder layers generates multi-modal impact tokens for the target object to generate the final prediction mask.
As shown in Figure~\ref{fig:main}, the input RGB image passes through the visual encoder to generate multi scale fused encoder feature $\mathbf{X_v} \in \mathbb{R} ^ {C \times H \times W}$, where $C$ is the number of embedding channels, and $H$ and $W$ are the height and width of the feature, respectively. For the text input, it goes through a pre-trained language encoder to create the language feature $\mathbf{X_t} \in \mathbb{R} ^ {C \times L}$, where $L$ is the length of language feature. This two features in different modalities are refined to tokens $\mathbf{T_v} \in \mathbb{R} ^ {C' \times \left( H \times W \right)}$ and $\mathbf{T_l} \in \mathbb{R} ^ {C' \times L}$, respectively. The generated $\mathbf{T_v}$ and $\mathbf{T_l}$ are used as inputs for two types of BTMAE: $BTMAE_{\left(l \rightarrow v \right)}$ and $BTMAE_{\left(v \rightarrow l \right)}$. In the case of $BTMAE_{\left(l \rightarrow v \right)}$, it generates a reconstructed visual token $\mathbf{T_v^r} \in \mathbb{R} ^ {C' \times \left( H \times W \right)}$ with refined spatial context information propagating from language to visual. Similarly, for $BTMAE_{\left(v \rightarrow l \right)}$, it generates a reconstructed language token $\mathbf{T_l^r} \in \mathbb{R} ^ {C' \times L}$ with enhanced linguistic context information propagating from visual to language. The process of $BTMAE$ is discussed in detail in Section~\ref{sec:3MAE}. Furthermore, ITA generates key impact tokens $\mathbf{T_v^i} \in \mathbb{R} ^ {C' \times K}$ on target objects from tokens reconstructed in each modality. The selected impact tokens are restored as prediction masks through impact token transformer (ITT) layers along with encoder feature $\mathbf{X_e} \in \mathbb{R} ^ {C \times H \times W}$.

\subsection{Bidirectional Token-Masking AutoEncoder}
\label{sec:3MAE}
Traditional MAE is based on a self-learning approach that erases the image in random patches and reconstructs it back to the original image. This approach performs well in enabling the model to learn rich visual contextual information to understand the whole image from the masked image. In RIS tasks, it is important for the RIS task to effectively extract the relational contextual information hidden between images and text. Therefore, we propose a novel BTMAE structure inspired by MAE, as shown in Figure~\ref{fig:3mae}. As mentioned in Section~\ref{intro}, to address the generalization problem in RIS tasks and to model cross-modal features, we divide the process of BTMAE into two steps. Step 1 involves both the BTMAE encoder and decoder to generate the auto-encoded image. Step 2 performs token refinement using only the BTMAE encoder. Figure~\ref{fig:3mae} (a) and (b) depict $BTMAE_{\left(l \rightarrow v \right)}$, but note that $BTMAE_{\left(v \rightarrow l \right)}$ also has the same model structure. Here are more specific description of our process. First, positional encoding and random masking are performed on the visual tokens $\mathbf{T_v}$ projected from the visual feature $\mathbf{X_v}$ to generate a masked token $\mathbf{T_v^m}$. In more detail, if the masking ratio is $\alpha$, the size of $\mathbf{T_v^m}$ is $C' \times \left( \lfloor H \times W \times \left( 1-\alpha \right) \rfloor \right)$. Second, $\mathbf{T_v^m}$ is forwarded to the $E$ transformer encoders as input feature. These BTMAE transformer encoders include multi-head attentions (MHA), layer normalization, and feedforward neural networks (FFN) as illustrated in Figure~\ref{fig:3mae}. Notably, a different modality tokens $\mathbf{T_l}$ from the query, is used as the key and value in the second MHA of each block. Third, we apply random masking tokens and positional encoding to the encoder output tokens in the same manner as MAE. These masking tokens are precisely inserted into the masked positions of $\mathbf{T_v}$. Consequently, the size of $\mathbf{T_v^m}$ becomes identical to that of $\mathbf{T_v}$. Finally, we pass these tokens through $D$ transformer decoder blocks and reshape and project them to create an auto-encoded image $\mathbf{I_{pred}}$. The decoder uses $\mathbf{T_l}$ as key and value the same as the BTMAE encoder. As mentioned above, step 2 generates enhanced visual tokens $\mathbf{T_v^r}$ using only unmasked $\mathbf{T_v}$, $\mathbf{T_l}$, and $E$ transformer encoders. Furthermore, $BTMAE_{\left(v \rightarrow l \right)}$ has the same structure as $BTMAE_{\left(l \rightarrow v \right)}$ and consequently auto-encodes unmasked language features.

BTMAE has the same visual transformer-based encoder and decoder structure as traditional MAE, but there are several differences: 1) While the traditional MAE performs random masking at the image patch level, BTMAE randomly masks the image tokens $\mathbf{T_v}$ generated from the backbone encoder. As explained earlier, we use this approach to generalize our model more effectively on the RIS dataset. We demonstrate in detail in Section~\ref{ablation} that this token-level masking is more effective than patch-level masking. 2) The encoder and decoder of MAE use self-attention based transformer blocks with only image features as key, query, and value. However, the proposed BTMAE employs queries and tokens $\mathbf{T_l}$ from a different modality as keys and values. Due to these structural differences, the proposed BTMAE can model multimodal correlations that traditional MAEs cannot. 3) MAE is trained during the pretraining phase and follows a two-stage strategy, where only the model encoder is used during the finetuning process. In contrast, BTMAE performs these tasks in a single training process. In other words, the steps 1 and 2 in Figure~\ref{fig:3mae} take place sequentially within a single training iteration. Figure~\ref{fig:3mae} (c) shows these process in clearly.

\subsection{Impact Token Attention}
From the enhanced multimodal tokens $\mathbf{T_v^r}$ and $\mathbf{T_l^r}$, the ITA generates impact tokens that play a key role in inferring a accurate pred mask. As shown in the Figure~\ref{fig:main}, the ITA generates impact tokens $\mathbf{T_v^i}$ by sampling tokens. The visual tokens $\mathbf{T_v^r}$ have rich local and global context information due to the encoder's patch-based attention structure. However, indiscriminate integration of this information during decoding may cause noise in mask generation, which hinders accurate mask generation. Therefore, we generate impact tokens $\mathbf{T_v^i}$ through spatial global average pooling to smooth $\mathbf{T_v^r}$. This process reduces the impact of noise from unnecessary tokens and consequently improves model performance. $\mathbf{T_v^r}$ and $\mathbf{T_l^r}$ pass through MHA and FFN layers and then reshapes it to the same size as $\mathbf{X_v}$. Next, for the separation and localization of target objects and background, it uses $1 \times 1$ convolution and pixel-wise softmax operations to generate spatial sampling masks $\mathbf{S_v} \in \mathbb{R} ^ {K \times H \times W}$, where $K$ is the number of visual impact tokens. Therefore, if we define the $k$-th channel of $\mathbf{S_v}$ as $\mathbf{S_{v^k}} \in \mathbb{R} ^ {1 \times H \times W}$, then this process is expressed as follows:

\begin{equation}
	\mathbf{S_{v^k \left(x,y\right)}} = \frac{e^{\mathbf{X^r_{v^k \left(x,y\right)}}}}{\sum_{k=1}^{K} e^{\mathbf{X^r_{v^k \left(x,y\right)}}}},
\end{equation}

\noindent
where $\left(x, y \right)$ are the pixel coordinates and $k=1,2,...,K$. $\mathbf{X^r_v} \in \mathbb{R} ^ {K \times H \times W}$ is the output of the standard self-attention sequences~\cite{vaswani2017attention}, as shown in Figure~\ref{fig:main}. Therefore, $\mathbf{X^r_v}$ is represented as follows:

\begin{equation}
	\begin{aligned}
		& \mathbf{X _ { v } ^ { r }} =f _ { 1 \times 1 } \left ( f _ { FFN } \left ( \mathbf{X _ { att }} \right ) \right ), \\
		& \mathbf{X _ { att }} =\psi \left ( \frac{ \mathbf{w _ { Q } T _ { v } ^ { r }} \left ( \mathbf{w _ { K } T _ { l } ^ { r }} \right ) ^ { \top } } { \sqrt { C ^ { ' } } } \right ) \left ( \mathbf{w _ { V } T _ { l } ^ { r }} \right ) + \mathbf{T _ { v } ^ { r }},
	\end{aligned}
\end{equation}

\noindent
where $\mathbf{w_K} \in \mathbb{R} ^ {C' \times C'}$, $\mathbf{w_Q} \in \mathbb{R} ^ {C' \times C'}$, and $\mathbf{w_V} \in \mathbb{R} ^ {C' \times C'}$ are learnable projection matrices of key, query, and value operations, respectively. Furthermore, $\psi \left(.\right)$ and $f \left(.\right)$ represent softmax and $1 \times 1$ convolution operations, separately. As a result, we generate $K$ impact tokens $\mathbf{T_v^i}=\left\{ \mathbf{T_{v^1}^i},\mathbf{T_{v^2}^i},...,\mathbf{T_{v^K}^i} \right\}$ through the following process:

\begin{equation}
	\mathbf{T_{v^k}^i} = \frac{\sum_{x=1}^{H} \sum_{y=1}^{W} \left(\mathbf{S_{{v^k} \left(x,y\right)}} \cdot \mathbf{X_{ \left(x,y\right)}}\right)}{\sum_{x=1}^{H} \sum_{y=1}^{W} \mathbf{S_{{v^k} \left(x,y\right)}}},
\end{equation}

\noindent
where $\mathbf{T_{v^k}^i}$ is the $k$-th visual impact token.

The last process of ITA involves fusing the multimodal impact tokens generated through the impact token transformer (ITT) with the encoder features to represent the area of interest. The structure of the proposed ITT is depicted in Figure~\ref{fig:3mae}. The ITT has a structure similar to a standard transformer decoder. The first MHA block fuses important tokens between the two modalities, while the second MHA block performs attention based on tokens from the encoder features $\mathbf{X_e}$ to capture the area of interest.

\begin{table*}
	\centering 
	\caption{Quantitative evaluation on the RefCOCO~\cite{yu2016modeling}, RefCOCO+~\cite{yu2016modeling}, and G-Ref~\cite{mao2016generation, nagaraja2016modeling}. The symbol * represents the results trained in the combination trainset of three datasets.}
	\vspace{-0.3cm}
	\scriptsize
	\begin{tabular}{c|l|lll|lll|ccc|ccc}
		\hline
		&            &             &            &        & \multicolumn{3}{c|}{RefCOCO~\cite{yu2016modeling}} & \multicolumn{3}{c|}{RefCOCO+~\cite{yu2016modeling}} & \multicolumn{3}{c}{G-Ref~\cite{mao2016generation, nagaraja2016modeling}} \\ \cline{6-14} 
		& Method     & Publication & Visual        & Language     & val     & test A   & test B  & val     & test A   & test B   & val U  & test U  & val G  \\ \hline
		\multirow{9}{*}{\rot{mIoU}}
		& VLT~\cite{ding2021vision}        & ICCV 2021    & Darknet-53 & Bi-GRU & 65.65   & 68.29    & 62.73   & 55.50    & 59.20     & 49.36    & 52.99  & 56.65   & 49.76  \\
		& RefTr~\cite{li2021referring}       & NeurIPS 2021   & ResNet-101 & BERT   & 74.34  & 76.77    & 70.87    & 66.75   & 70.58    & 59.40    & 66.63  & 67.39   & -      \\
		& CRIS~\cite{wang2022cris}       & CVPR 2022   & ResNet-101 & CLIP   & 70.47   & 73.18    & 66.10   & 62.27   & 68.08    & 53.68    & 59.87  & 60.36   & -      \\
		& LAVT~\cite{yang2022lavt}       & CVPR 2022   & Swin-B     & BERT   & 72.73   & 75.82    & 68.79   & 62.14   & 68.38    & 55.10    & 61.24  & 62.09   & 60.50  \\
		& SeqTR~\cite{zhu2022seqtr}       & ECCV 2022   & Darknet-53 & Bi-GRU   & 71.70   & 73.31    & 69.82   & 63.04   & 66.73    & 58.97    & 64.69  & 65.74   & -      \\
		& PolyFormer*~\cite{liu2023polyformer} & CVPR 2023   & Swin-B     & BERT   & 75.96  & 77.09    & 73.22   & \underline{70.65}   & \underline{74.51}    & \underline{64.64}    & 69.36  & \underline{69.88}   & -      \\
		& CGFormer~\cite{tang2023contrastive}   & CVPR 2023   & Swin-B     & BERT   & 76.93   & 78.70    & 73.32   & 68.56   & 73.76    & 61.72    & 67.57  & 67.83   & 65.79  \\ \cline{2-14} 
		& Ours       &             & ResNet-101 & BERT   & 74.56   & 76.97 & 71.76  & 67.25 & 71.13 & 60.00 & 67.13 & 67.87  & 65.11      \\
		& Ours       &             & DarkNet-53 & BERT   & 74.91 & 77.03 & 71.72 & 67.03 & 71.14 & 60.65 & 67.34 & 67.93 & 65.45 \\
		& Ours       &             & Swin-B     & BERT   & \underline{77.21} &  \underline{79.00}  &  \underline{74.15}  &  69.38  &  74.25  & 62.54  &  \underline{69.49}      &  69.52       & \underline{67.36}       \\
		& Ours*       &             & Swin-B     & BERT   & \textbf{77.46} &  \textbf{79.25}  &  \textbf{74.22}  &  \textbf{70.81}  &  \textbf{74.71}  & \textbf{64.84}  & \textbf{69.65}   & \textbf{70.01}   & \textbf{67.87} \\ \hline
		\multirow{11}{*}{\rot{oIoU}} 
		& LSCM~\cite{hui2020linguistic}       & ECCV 2020  & ResNet-101 & LSTM & 61.47   & 64.99    & 59.55   & 49.34   & 53.12    & 43.50    & -  & -   & -  \\
		& CGAN~\cite{luo2020cascade}       & ACMMM 2020  & DarkNet-53 & Bi-GRU & 64.86   & 68.04    & 62.07   & 51.03   & 55.51    & 44.06    & 51.01  & 51.69   & 46.54  \\
		& LTS~\cite{jing2021locate}        & CVPR 2021   & DarkNet-53 & Bi-GRU & 65.43   & 67.76    & 63.08   & 54.21   & 58.32    & 48.02    & 54.40  & 54.25   & -      \\
		& ReSTR~\cite{kim2022restr}      & CVPR 2022   & ViT-B-16   & GloVe  & 67.22   & 69.30    & 64.45   & 55.78   & 60.44    & 48.27    & -      & -       & -      \\
		& SLViT~\cite{ouyang23slvit}      & IJCAI 2023  & SegNeXt    & BERT   & 74.02   & 76.91    & 70.62   & 64.07   & 69.28    & 56.14    & 62.75  & 63.57   & 60.94  \\
		& M3Att~\cite{liu2023multi}      & TIP 2023    & Swin-B     & BERT   & 72.97   & 76.23    & 70.36   & 65.34   & 70.50    & 56.98    & 64.92  & 67.37   & 63.90  \\
		& SADLR~\cite{yang2023semantics}      & AAAI 2023   & Swin-B     & BERT   & 74.24   & 76.25    & 70.06   & 64.28   & 69.09    & 55.19    & 63.60  & 63.56   & 61.16  \\
		& MCRES~\cite{xu2023meta}      & CVPR 2023   & Swin-B     & BERT   & 74.92   & 76.98    & 70.84   & 64.32   & 69.68    & 56.64    & 63.51  & 64.90   & 61.63  \\
		& PolyFormer*~\cite{liu2023polyformer} & CVPR 2023   & Swin-B     & BERT   & 74.82   & 76.64   & \underline{71.06}   & \underline{67.64}   & \textbf{72.89}    & 59.33    & \underline{67.76}  & \underline{69.05}   & -      \\
		& CGFormer~\cite{tang2023contrastive}   & CVPR 2023   & Swin-B     & BERT   & 74.75   & 77.30     & 70.64   & 64.54   & 71.00     & 57.14    & 64.68  & 65.09   & 62.51  \\
		& ETRIS~\cite{xu2023bridging}      & ICCV 2023   & ViT-B-16   & CLIP   & 70.51   & 73.51    & 66.63   & 60.10   & 66.89    & 50.17    & 59.82  & 59.91   & 57.88  \\
		& DMMI~\cite{hu2023beyond}       & ICCV 2023   & Swin-B     & BERT   & 74.13   & 77.13    & 70.16   & 63.98   & 69.73    & 57.03    & 63.46  & 64.19   & 61.98  \\ \cline{2-14} 
		& Ours       &             & ResNet-101 & BERT   & 72.45   & 75.83    & 70.00   & 65.12   & 70.01    & 56.53    & 63.93  & 65.34   & 61.27        \\
		& Ours       &             & DarkNet-53 & BERT   & 72.53   & 75.99    & 69.87   & 65.02   & 70.05    & 56.83    & 64.12  & 65.51   & 61.80        \\
		& Ours       &             & Swin-B     & BERT    & \underline{74.95}   & \underline{77.41}     & 70.89   & 66.21   & 71.32       & \underline{59.56}    & 66.72  & 67.57   & \underline{64.00} \\
		& Ours*       &             & Swin-B     & BERT   & \textbf{75.01}  & \textbf{77.53}    &  \textbf{71.22}  & \textbf{67.70}  &  \underline{72.85}   & \textbf{59.92}  & \textbf{68.29} & \textbf{69.35}       & \textbf{66.03}    \\ \hline
	\end{tabular}
	\label{Table:results}
\end{table*}

\subsection{Objective Function}
First, we use mean squared error loss in $BTMAE_{\left(l \rightarrow v \right)}$ to reconstruct the input image the same as MAE. This process can be expressed as follows:

\begin{equation}
	\mathcal{L}_{BTMAE_{\left(l \rightarrow v \right)}}=\sum_{x,y}^{}\left(\mathbf{I_{Pred}} \left(x,y\right) - \mathbf{I_{GT}} \left(x,y\right)\right)^2,
\end{equation}

\noindent
where $\mathbf{I_{Pred}}$ and $\mathbf{I_{GT}}$ refer to the predicted RGB image and the original RGB image respectively. Second, For $BTMAE_{\left(v \rightarrow l \right)}$, which reconstructs language features, the cross-entropy loss $\mathcal{L}_{BTMAE_{\left(v \rightarrow l \right)}}$ between predicted language tokens $\mathbf{T_l^r}$ and original language tokens $\mathbf{X_l}$ is calculated in the same way proposed by BERT~\cite{devlin2018bert}. Third, we use sum of IOU loss and weighted binary cross-entropy loss to assign more weight to the hard case pixels. we define these loss functions as follows:

\begin{equation}
	\begin{aligned}
		& \mathcal{L}_{I O U}=1-\frac{\sum_{\left(x,y\right)} \min \left(\mathbf{M}_{\mathbf{P}}, \mathbf{M}_{\mathbf{G}}\right)}{\sum_{\left(x,y\right)} \max \left(\mathbf{M}_{\mathbf{P}}, \mathbf{M}_{\mathbf{G}}\right)}, \\
		& \mathcal{L}_{b c e}^w=-\sum_{\left(x,y\right)}w\left[\mathbf{M}_{\mathbf{G}} \ln \left(\mathbf{M}_{\mathbf{P}}\right)+\left(1-\mathbf{M}_{\mathbf{GT}}\right) \ln \left(1-\mathbf{M}_{\mathbf{P}}\right)\right],
	\end{aligned}
\end{equation}

\noindent
where $w=\sigma\left|\mathbf{M_P}-\mathbf{M_G}\right|$ and $\left(x,y\right)$ is pixel coordinate. Also $\mathbf{M_G}$ and $\mathbf{M_P}$ are ground truth maps and prediction maps, respectively. As a result, the total loss function is expressed as follows: 

\begin{equation}
	\mathcal{L}_{\text {total}}=\mathcal{L}_{BTMAE_{\left(l \rightarrow v \right)}} + \mathcal{L}_{BTMAE_{\left(v \rightarrow l \right)}} + \mathcal{L}_{I O U}+\mathcal{L}_{b c e}^w.
\end{equation}

\section{Experiments}
\subsection{Datasets \& Evaluation Metrics}
In this research, we performed experiments using three commonly used benchmark datasets, namely RefCOCO~\cite{yu2016modeling}, RefCOCO+~\cite{yu2016modeling}, and GRef~\cite{mao2016generation, nagaraja2016modeling}. While these datasets are all based on MSCOCO~\cite{lin2014microsoft}, they come with distinct annotation settings. RefCOCO~\cite{yu2016modeling}, for instance, features relatively short descriptions for 50,000 objects, averaging around 3.5 words. RefCOCO+~\cite{yu2016modeling} is more challenging, since it does not describing absolute locations of referents for 49,856 objects. In contrast, G-Ref~\cite{mao2016generation, nagaraja2016modeling} has longer expressions with an average word count of 8.4 for 54,822 objects. To maintain consistency with prior research~\cite{ding2021vision, wang2022cris, yang2022lavt}, we divided RefCOCO~\cite{yu2016modeling} and RefCOCO+~\cite{yu2016modeling} into training, validation, test A, and test B subsets. For G-Ref, we utilized both UMD and Google partitions for evaluation.

Since previous works use either mean intersection-over-union (mIoU) or overall intersection-over-union (oIoU) as an evaluation metric, we evaluate our model using both metrics for a fair comparison. The oIoU metric quantifies the proportion of the collective intersection area compared to the collective union area derived from all test samples. On the other hand, the mIoU computes the mean IoU score across all individual samples within the entire test dataset.

\subsection{Implementation Details}
In this paper, we use ResNet-101~\cite{he2016deep}, Darknet-53~\cite{redmon2018yolov3}, and Swin-B~\cite{liu2021swin} pretrained on ImageNet~\cite{deng2009imagenet} as the backbone visual endoder for a fair comparison with previous works. Also, as a backbone language encoder we use an uncased version of BERT model~\cite{devlin2018bert} with 12 layers and the hidden size of 768, and weights initialized from pretrained weight provided by HuggingFace~\cite{wolf2020transformers}, and image size is $480 \times 480$. We set the number of encoder layers as $E=4$ and the number of decoder layers as $D=4$ in the proposed $BTMAE_{\left(l \rightarrow v \right)}$ and $BTMAE_{\left(v \rightarrow l \right)}$ modules. Furthermore, we experimentally set the token masking ratio $\alpha$ of both BTMAE to 0.5. Comparative experiments on various masking ratios are covered in Section~\ref{ablation}. Finally, we reduce the number of remaining tokens for each ITA layer by setting the number of impact tokens $\mathbf{T_v^r}$ generated by ITA to 16, 8, and 2. As a result, in the last ITA layer, only two tokens remain, corresponding to the foreground and background. The Adam optimizer~\cite{kingma2014adam} is used for network training and fine-tuning with hyperparameters $\beta_1 = 0.9$, $\beta_2 = 0.999$, and $\epsilon = 10^{-8}$. The learning rate decreases from $10^{-4}$ to $10^{-5}$ using a cosine annealing scheduler~\cite{loshchilov2016sgdr}. The total number of epochs is set to 200, with a batch size of 12. The experiments are conducted on a two NVIDIA RTX A6000 GPUs and are implemented using the PyTorch deep-learning framework.

\begin{figure*}[t]
	\setlength{\belowcaptionskip}{-24pt}
	\begin{center}
		\includegraphics[width=1\linewidth]{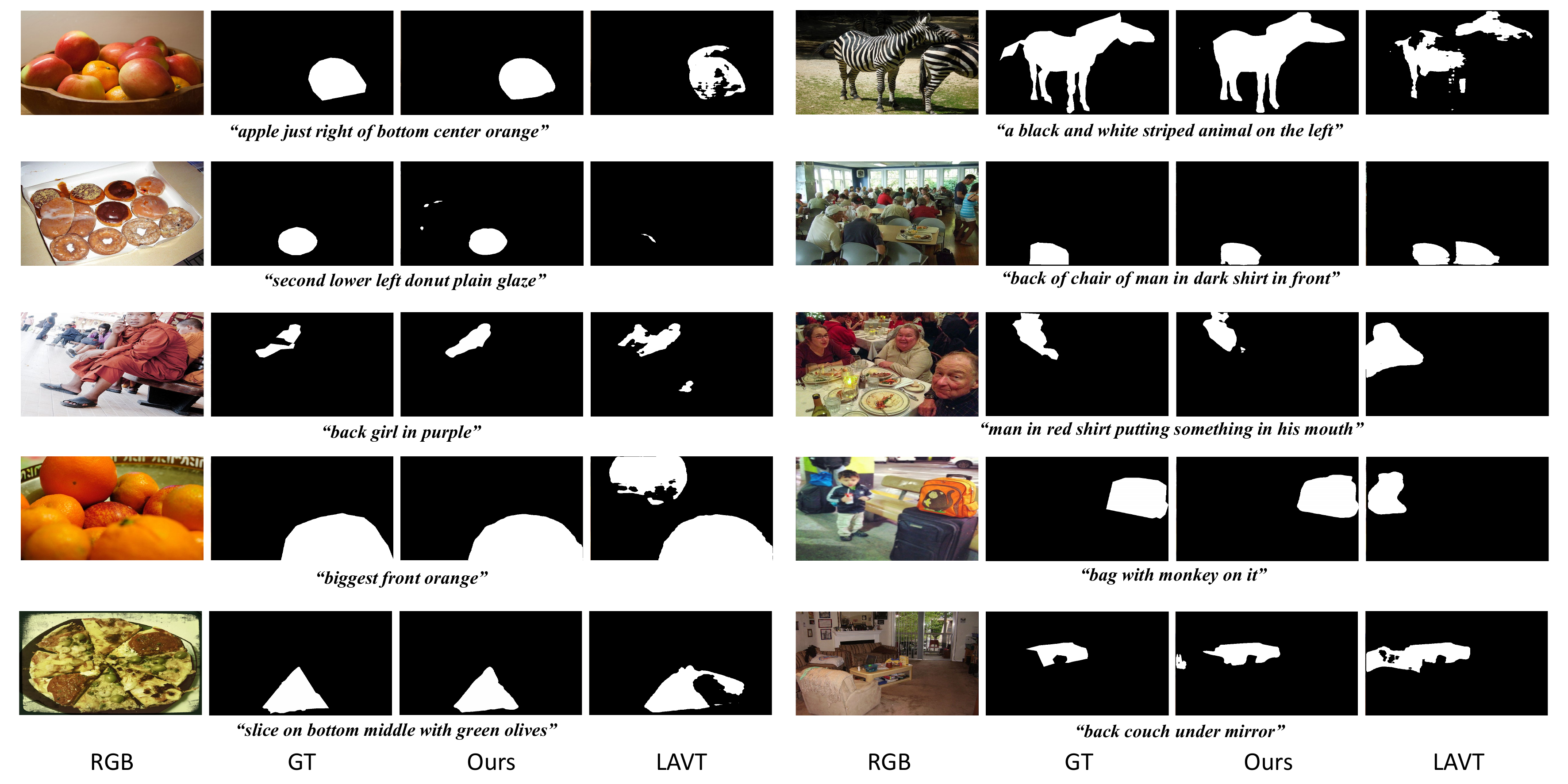}
		\vspace{-0.7cm}
		\caption{Comparisons of prediction masks generated by LAVT~\cite{yang2022lavt} and our BTMAE.}
		\label{fig:result}
	\end{center}
\end{figure*}

\subsection{Results}
\noindent
\textbf{Quantitative results.} Table~\ref{Table:results} presents a quantitative comparison between state-of-the-art fully-supervised methods~\cite{ding2021vision, li2021referring, wang2022cris, yang2022lavt, zhu2022seqtr, liu2023polyformer, tang2023contrastive, hui2020linguistic, luo2020cascade, jing2021locate, kim2022restr, ouyang23slvit, liu2023multi, yang2023semantics, xu2023meta, xu2023bridging, hu2023beyond} and the proposed model on three benchmarks. Since existing methods utilize various visual encoders, we evaluate our model using the most commonly used ResNet-101~\cite{he2016deep}, Darknet-53~\cite{redmon2018yolov3}, and Swin-B~\cite{liu2021swin} encoders as backbones to ensure a fair comparison. For the language encoder, we employ the widely used BERT~\cite{devlin2018bert}. We also attach results for other language models such as CLIP~\cite{radford2021learning} in the supplementary material. As shown in Table~\ref{Table:results}, the proposed BTMAE outperforms the existing methods on all benchmark datasets and evaluation metrics. In particular, the performance comparison with models using the same visual and language backbone as BTMAE supports that the proposed model can effectively model visual-language relations. Additionally, BTMAE maintains robust performance even on the G-Ref~\cite{mao2016generation, nagaraja2016modeling} dataset consisting of more than 8 words on average, which shows that the proposed model has good generalization performance.

\begin{figure}[t]
	\setlength{\belowcaptionskip}{-24pt}
	\begin{center}
		\includegraphics[width=\linewidth]{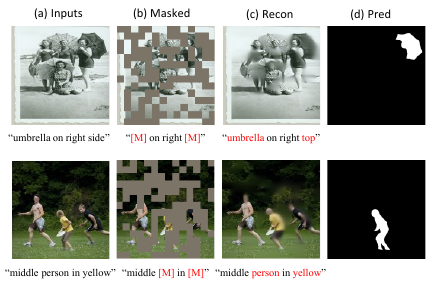}
		\vspace{-0.7cm}
		\caption{Visualizing the autoencoding results of BTMAE. (a) Input image and text description. (b) Token masked image and text. (c) Reconstructed image and text. (d) Model final prediction mask. Note that since we are masking at the token level, (b) visualizes the positions of the masked tokens rather than the actual input.}
		\label{fig:recon}
	\end{center}
\end{figure}

\noindent
\textbf{Qualitative results.} Figure~\ref{fig:result} shows the prediction mask visualization results of BTMAE under various challenging situations. We also add the results of LAVT~\cite{yang2022lavt}, a representative attention-based visual-language fusion model. Note that, as shown in Table~\ref{Table:results}, LAVT~\cite{yang2022lavt} uses the same visual and language backbone as our model. As shown in Figure~\ref{fig:result}, the proposed BTMAE demonstrates robust target object prediction in challenging situations such as a large number of common objects. Furthermore, the BTMAE is capable of modeling more complex relationships between images and texts, resulting in encouraging performance on complex sentences.

\begin{table}
	\centering 
	\caption{mIoU performance with different combinations of our contributions. If the proposed contribution is not applied to the encoder or decoder, we apply a simple attention-based encoder or decoder proposed by LAVT~\cite{yang2022lavt}.}
	\vspace{-0.3cm}
	\resizebox{\columnwidth}{!}{
		\begin{tabular}{c|ccc|c|c|c}
			\hline
			\multirow{2}{*}{Index} & \multicolumn{3}{c|}{Method} & RefCOCO~\cite{yu2016modeling} & RefCOCO+~\cite{yu2016modeling} & G-Ref~\cite{mao2016generation, nagaraja2016modeling}  \\ \cline{2-7} 
			& $BTMAE_{\left(l \rightarrow v \right)}$ & $BTMAE_{\left(v \rightarrow l \right)}$ & ITA     & test A   & test A    & test-U \\ \hline
			(a)                    &                          &                          &         & 75.82  & 68.38 & 62.09 \\
			(b)                    & \ding{51}                 &                          &         & 77.35 & 71.25 & 64.78 \\
			(c)                    &                          & \ding{51}                 &         &  76.84  & 70.93  & 63.41 \\
			(d)                    & \ding{51}                 & \ding{51}                 &         &  78.07 & 73.55 & 68.34 \\
			(e)                    &                          &                          & \ding{51} &  76.81  & 70.77 & 64.12  \\
			(f)                    & \ding{51}                 & \ding{51}                 & \ding{51} & \textbf{79.25} & \textbf{74.71}  & \textbf{70.01} \\ \hline
		\end{tabular}
	}
	\vspace{-0.3cm}
	\label{Table:combination}
\end{table}

\begin{table}
	\centering 
	\caption{Comparison of generalization oIoU performance under different masking and training methods. The trainset uses RefCOCO~\cite{yu2016modeling} and the testset uses testA. When the masking method is patch, it has the same structure as the traditional MAE, and for the two-stage case, it follows the traditional pretrain followed by finetune. If the number of stages is one, we follow the training scheme of the proposed BTMAE.}
	\vspace{-0.3cm}
	\resizebox{\columnwidth}{!}{
		\begin{tabular}{cc|ccccccccc}
			\begin{tabular}{cc|ccc|ccc|ccc}
				\hline
				\multicolumn{2}{c|}{Index}                           & (a)                  & (b)                  & (c)                   & (d)                  & (e)                  & (f)                   & (g)                  & (h)                  & (i)                  \\ \hline
				\multicolumn{2}{c|}{Dataset Size}                    & 10K                  & 10K                  & 10K                   & 20K                  & 30K                  & 30K                   & 50K                  & 50K                  & 50K                  \\ \hline
				\multicolumn{2}{c|}{Stage}                           & 2                    & 2                    & 1                     & 2                    & 2                    & 1                     & 2                    & 2                    & 1                    \\ \hline
				\multicolumn{1}{c|}{\multirow{2}{*}{Method}} & Patch & \ding{51} &                      &                       & \ding{51}  &                      &                       & \ding{51} &                      &                      \\ \cline{2-2}
				\multicolumn{1}{c|}{}                        & Token &                      & \ding{51} & \ding{51} &                      & \ding{51} & \ding{51} &                      & \ding{51} & \ding{51} \\ \hline
				\multicolumn{2}{l|}{RefCOCO test A}                  & \multicolumn{1}{l}{74.64} & \multicolumn{1}{l}{76.16} & \multicolumn{1}{l|}{76.22} & \multicolumn{1}{l}{77.17} & \multicolumn{1}{l}{78.36} & \multicolumn{1}{l|}{78.24} & \multicolumn{1}{l}{78.32} & \multicolumn{1}{l}{79.02} & \multicolumn{1}{l}{79.00} \\ \hline
			\end{tabular}
		\end{tabular}
	}
	\label{Table:mae}
\end{table}

\noindent
\textbf{Visual \& language reconstruction results.}
Figure~\ref{fig:recon} shows the results of the reconstructed images and text generated from the BTMAE. Unlike the traditional MAE, the BTMAE masks the input in tokenized form without patch-wise masking. More specifically, for better visualization, we indicate the positions of masked tokens before tokenization, as shown in (b). Although the reconstructed text is somewhat different from the ground truth, it can be seen that the proposed model effectively models multimodal relationships in that it is harmoniously connected to the image.

\begin{figure}[t]
	\setlength{\belowcaptionskip}{-24pt}
	\begin{center}
		\includegraphics[width=\linewidth]{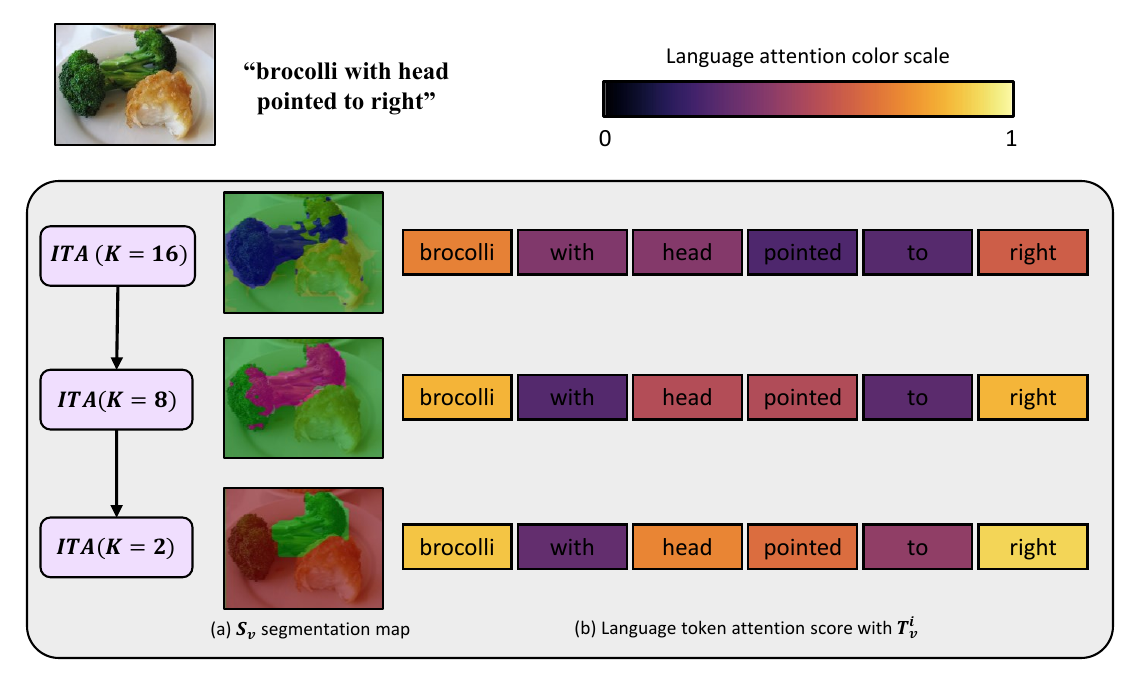}
		\vspace{-0.7cm}
		\caption{Visual and language domain visualization maps based on impact tokens that are progressively selected as they pass through the ITA decoder layer. (a) $\mathbf{S_v}$ segmentation map for each token, which is the area where impact tokens $\mathbf{T_v^i}$ are selected, and (b) visualization of the relative similarity scores of language tokens and impact tokens. The proposed ITA layer samples impact tokens while gradually removing noise about objects other than the target and focusing on more important text information.}
		\label{fig:ablation}
	\end{center}
\end{figure}

\subsection{Ablation Analysis}
\label{ablation}
In this section, we demonstrate the effectiveness of the proposed model through various comparative experiments. All experiments are conducted using the Swin-B~\cite{liu2021swin} visual encoder and the BERT~\cite{devlin2018bert} language encoder.

\noindent
\textbf{Effect of BTMAE.} Index (b), (c), and (d) in Table~\ref{Table:combination} and the results in Table~\ref{Table:mae} show the effectiveness of the proposed BTMAE. As shown in Table~\ref{Table:combination}, the use of BTMAE leads to significant performance improvements on all datasets, especially the best performance when $BTMAE_{\left(l \rightarrow v \right)}$ and $BTMAE_{\left(v \rightarrow l \right)}$ are used together. This shows that the proposed BTMAE model is capable of modeling the strong relationship between vision and language. Furthermore, we demonstrate the generalization performance of BTMAE on the RIS task in Table~\ref{Table:mae}. We subsample the RefCOCO~\cite{yu2016modeling} train dataset to 10K, 30K, and 50K and compare the performance of BTMAE based on token masking with traditional MAE based on patch masking. As shown in this table, we can see that the proposed BTMAE maintains prediction performance on smaller datasets compared to the traditional MAE. This supports that BTMAE has better generalization performance compared to the traditional MAE. In addition, the training method is also effective, as the BTMAE shows almost the same performance compared to the two-step training process.

\noindent
\textbf{Effect of ITA.} Index (e), (f) of Table~\ref{Table:combination} and Figure~\ref{fig:ablation} show the effectiveness of the proposed ITA. As shown in the table, ITA achieves significant performance gains on all datasets, especially when combined with BTMAE. Furthermore, we visualize the effect of the impact token sampling method as shown in Figure~\ref{fig:ablation}. As described earlier, the proposed ITA decoder consists of a total of three layers and progressively samples 16, 8, and 2 impact tokens $\mathbf{T_v^i}$. In Figure~\ref{fig:ablation}, (a) shows the inference result of $\mathbf{S_v}$, which contains a spatial information mask from which impact tokens are selected, as a segmentation mask. Each channel of Sv is formed by a different colored segmentation map. As shown in this figure, the impact tokens are trained to progressively distinguish the target object from the background as they pass through the ITA. In addition, in (b) we visualize the similarity-based attentional scores between the impact tokens generated by each ITA layer and the input language tokens. We can see that the language feature also gives high attentional value to important words as the impact tokens are refined.

\begin{figure}[t]
	\setlength{\belowcaptionskip}{-24pt}
	\begin{center}
		\includegraphics[width=\linewidth]{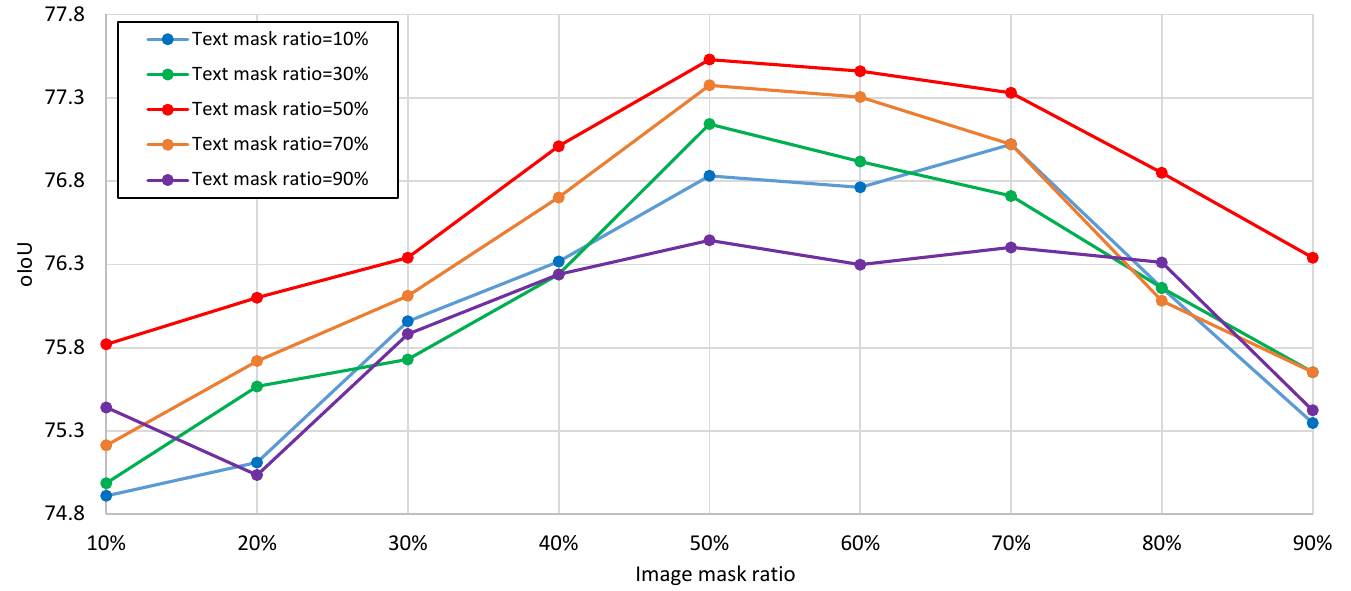}
		\vspace{-0.7cm}
		\caption{Comparison of oIoU performance in RefCOCO test A for different image token and text token mask ratios of the proposed BTMAE.}
		\label{fig:ratio}
	\end{center}
\end{figure}

\noindent
\textbf{Effect of masking ratio.} Figure~\ref{fig:ratio} compares the oIoU performance on the RefCOCO test set A for each modality of the proposed BTMAE based on token mask ratios. To apply various text token mask ratios, we use the three integrated datasets of RefCOCO, RefCOCO+, and G-Ref as training datasets, following~\cite{liu2023polyformer}. The best performance is observed when the image  and text mask token ratio are both 50\%, as shown in Figure~\ref{fig:ratio}. Generally, traditional MAE in the image domain is known to be most effective when masking around 70\% of the patches~\cite{he2022masked}. However, the proposed BTMAE aims to reconstruct the modeling of relationships between cross-modalities, rather than reconstructing the entire image from sparse single-modal information. For this reason, we believe that the optimal masking ratios for MAE and BTMAE are different.

\section{Conclusion}
Our proposed BTMAE addresses the limitations of existing RIS models by introducing a novel approach that enhances the understanding of complex and ambiguous contextual information in both language and image modalities. The effectiveness of our proposed method is validated through extensive experimentation on three widely used datasets, where the BTMAE achieves state-of-the-art performance. The model's ability to generate contextual tokens and mutually complement features across images and language significantly improves the robustness of RIS in handling challenging scenarios.
% Through comprehensive ablation studies, we demonstrate the significance of our approach in overcoming the limitations of existing models, making it a promising contribution to the evolution of RIS methodologies.

{
    \small
    \bibliographystyle{ieeenat_fullname}
    \bibliography{main}
}

% WARNING: do not forget to delete the supplementary pages from your submission 
% \input{sec/X_suppl}

\end{document}